\documentclass[letterpaper, 10 pt, conference]{ieeeconf}  

\usepackage{array}
\usepackage{arydshln}
\usepackage{multirow}
\usepackage{mathtools}
\usepackage{algorithm,algpseudocode}
\usepackage{ragged2e}
\usepackage{dsfont}

\usepackage{subcaption}
\usepackage[T1]{fontenc}
\usepackage{graphicx}
\usepackage{booktabs}
\usepackage{caption}
\IEEEoverridecommandlockouts 

\newcommand{\midsepremove}{\aboverulesep = 0mm \belowrulesep = 0mm}
\midsepremove
\newcommand{\midsepdefault}{\aboverulesep = 0.605mm \belowrulesep = 0.984mm}
\midsepdefault
    
\usepackage{graphics} 
\usepackage{amsmath} 
\usepackage{amssymb}  
\usepackage{mathtools}

\usepackage{array}
\newcolumntype{P}[1]{>{\centering\arraybackslash}p{#1}}
\title{\LARGE \bf
Pragmatic Embodied Spoken Instruction Following in Human-Robot Collaboration with Theory of Mind
}

\author{Lance Ying$^{1,2}$, Xinyi Li$^{3}$, Shivam Aarya$^{3}$, Yizirui Fang$^{3}$, Jason Xinyu Liu$^{4}$, Yifan Yin$^{3}$\\ Stefanie Tellex$^{4}$, Joshua B. Tenenbaum$^{1}$, Tianmin Shu$^{3}$
\thanks{$^{1}$Massachusetts Institute of Technology, Cambridge, MA 01239, USA
    {\tt\small \{lanceying, jbt\}@mit.edu}}%
\thanks{$^{2}$Harvard University, Cambridge, MA 02138, USA
    {\tt\small lanceying@seas.harvard.edu}}%
\thanks{$^{3}$Johns Hopkins University, Baltimore, MD 21218, USA
    {\tt\small \{xli383, saarya1,yfang52, yyin34, tianmin.shu\}@jhu.edu}}
\thanks{$^{4}$Brown University, Providence, RI 02912, USA
    {\tt\small \{xinyu\_liu, stefanie90\}@brown.edu}}}

\begin{document}

\maketitle
\thispagestyle{empty}
\pagestyle{empty}

\begin{abstract}
Spoken language instructions are ubiquitous in agent collaboration. However, in real-world human-robot collaboration, following human spoken instructions can be challenging due to various speaker and environmental factors, such as background noise or mispronunciation. When faced with noisy auditory inputs, humans can leverage the collaborative context in the embodied environment to interpret noisy spoken instructions and take pragmatic assistive actions. In this paper, we present a cognitively inspired neurosymbolic model, Spoken Instruction Following through Theory of Mind (SIFToM), which leverages a Vision-Language Model with model-based mental inference to enable robots to pragmatically follow human instructions under diverse speech conditions. We test SIFToM in both simulated environments (VirtualHome) and real-world human-robot collaborative settings with human evaluations. Results show that SIFToM can significantly improve the performance of a lightweight base VLM (Gemini 2.5 Flash), outperforming state-of-the-art VLMs (Gemini 2.5 Pro) and approaching human-level accuracy on challenging spoken instruction following tasks.

\end{abstract}

\section{Introduction}


Spoken language is the cornerstone of human cooperation, enabling the seamless coordination of actions and the communication of intent \cite{tomasello2010origins}. From a very young age, humans develop a remarkable proficiency for giving and following verbal instructions, a skill that is fundamental to collaborative activities. The recent proliferation of powerful Vision-Language Models (VLMs) has opened promising avenues for creating robotic agents that can collaborate with humans in a similarly natural fashion, integrating both visual and linguistic inputs across a wide range of applications \cite{ahn2022can, gao2024physically}. However, despite significant progress in controlled settings, a formidable challenge remains in transitioning these systems to the complexities of the real world.

\begin{figure}[t!]
    \centering
    \includegraphics[width=0.9\linewidth]{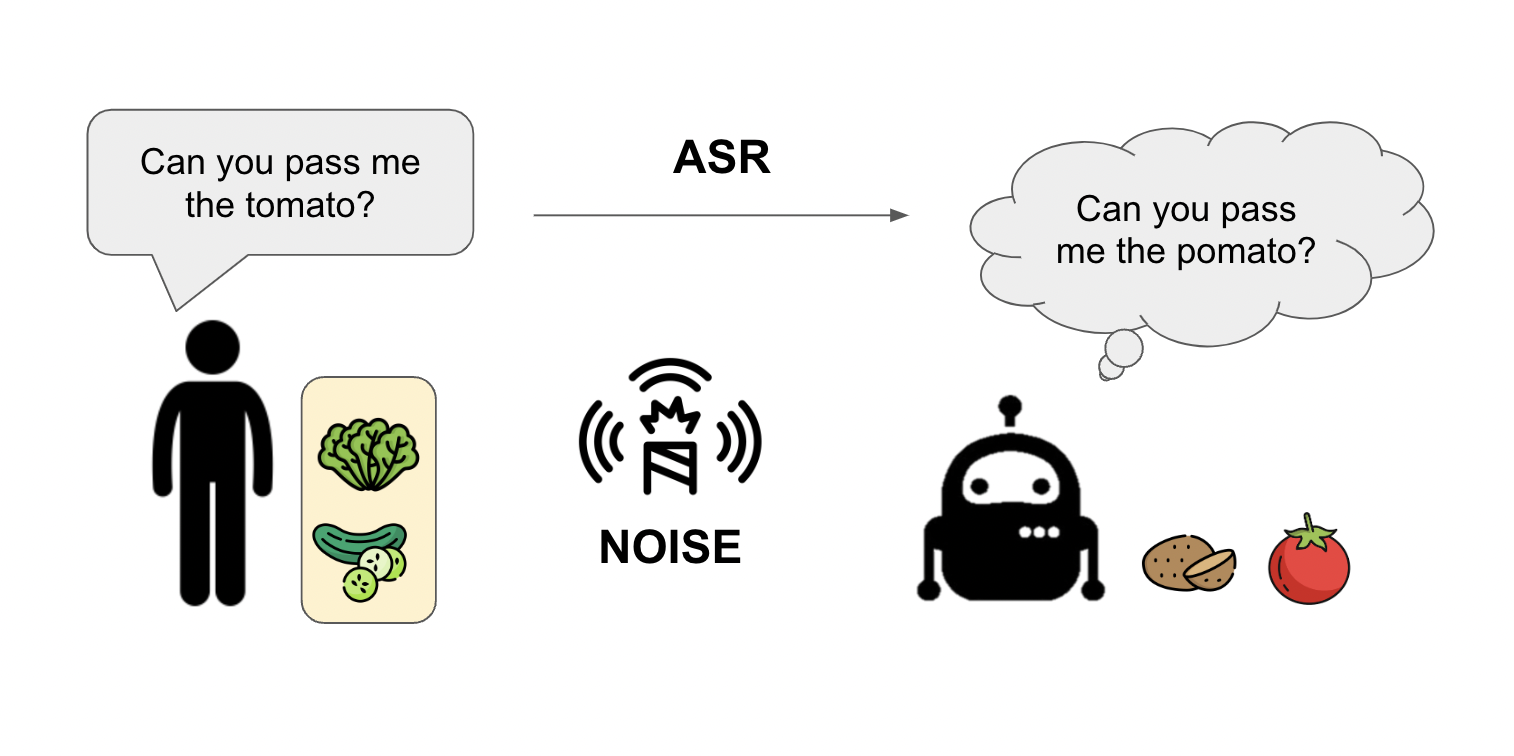}
    \caption{Example scenario where the human is asking for the tomato to make a salad. State-of-the-art automatic speech recognition (ASR) models or Vision-Language Models often cannot decode whether the human said tomato or potato due to noise or mispronunciation. However, SIFToM can make a pragmatic inference that the human is making a salad and is likely asking for a tomato.}
    \label{fig:intro}
    \vspace{-10pt}
\end{figure}

The crux of this challenge lies in the inherent fragility of human speech. In realistic, open-world environments—such as busy factories, dynamic mining sites, or even a typical home kitchen—spoken instructions are frequently corrupted by acoustic noise, speaker-specific idiosyncrasies such as accents and disfluencies. For a machine, these imperfections can render an instruction unintelligible, leading to interaction failures. Imagine the scenario depicted in Figure~\ref{fig:intro}, where speech recognition may give the robot the wrong transcription (potato instead of tomato) due to either noise in the speech or mispronunciation by the user. This wrong inference would result in completely failed assistance, even though there is only a single incorrect word. In stark contrast, humans adeptly overcome such ambiguity by leveraging a shared collaborative context. Rather than merely transcribing noisy auditory signals, we engage in pragmatic reasoning, inferring a speaker's underlying goals and intentions to disambiguate their words. This ability, often associated with a ``Theory of Mind,'' allows us to understand what was meant, not just what was said.

Inspired by this human capacity, we present a novel inference-time neurosymbolic framework, Spoken Instruction Following through Theory of Mind (SIFToM), for robust instruction following in human-robot collaboration. Our framework is designed to ground the interpretation of ambiguous spoken instructions within a formal model of mental inference. By observing a human collaborator's actions within a collaborative environment, SIFToM prompts a VLM to construct an environment and agent model by parsing multimodal inputs into structured representations, including scene graphs, action sequences, noisy language instructions, and goal space. This VLM-based model construction supports model-based reasoning, which extends a robot's capability of robust context-driven instruction following to pragmatically deduce the most likely intended human command in challenging human-robot collaboration scenarios.

We conducted a rigorous evaluation of SIFToM in both simulated environments and open-ended real-world human-robot collaborative settings with human subjects. For testing the model in controlled, simulated environments, we create a new speech instruction dataset, UnclearInstruct, which includes videos simulated in a realistic multiagent household simulator, VirtualHome 2 \cite{puig2020watch}, paired with real human spoken instructions injected with real-world noise. We also evaluated SIFToM for robots during real-world user assistance in a real kitchen environment when embedded in a Stretch robot. In both experiments, we show that SIFToM leveraging a lightweight base VLM (Gemini 2.5 Flash) can outperform a larger, state-of-the-art VLM (Gemini 2.5 Pro). Critically, our model's ability to correctly infer and act upon challenging spoken instructions begins to approach human-level accuracy, highlighting the substantial benefits of integrating pragmatic reasoning into robotic instruction following.

In sum, our main contribution includes (1) a novel neurosymbolic framework, Spoken Instruction Following
through Theory of Mind (SIFToM), for robust embodied assistance with unclear human spoken instructions; (2) a new spoken instruction dataset, UnclearInstruct, paired with an embodied assistance testbed for simulated spoken instruction following; (3) a real-world robot evaluation, comparing our neurosymbolic framework against state-of-the-art end-to-end VLM baselines on driving a robot assistant to help real humans.

\begin{figure*}[ht!]
    \centering
    \includegraphics[width=\linewidth]{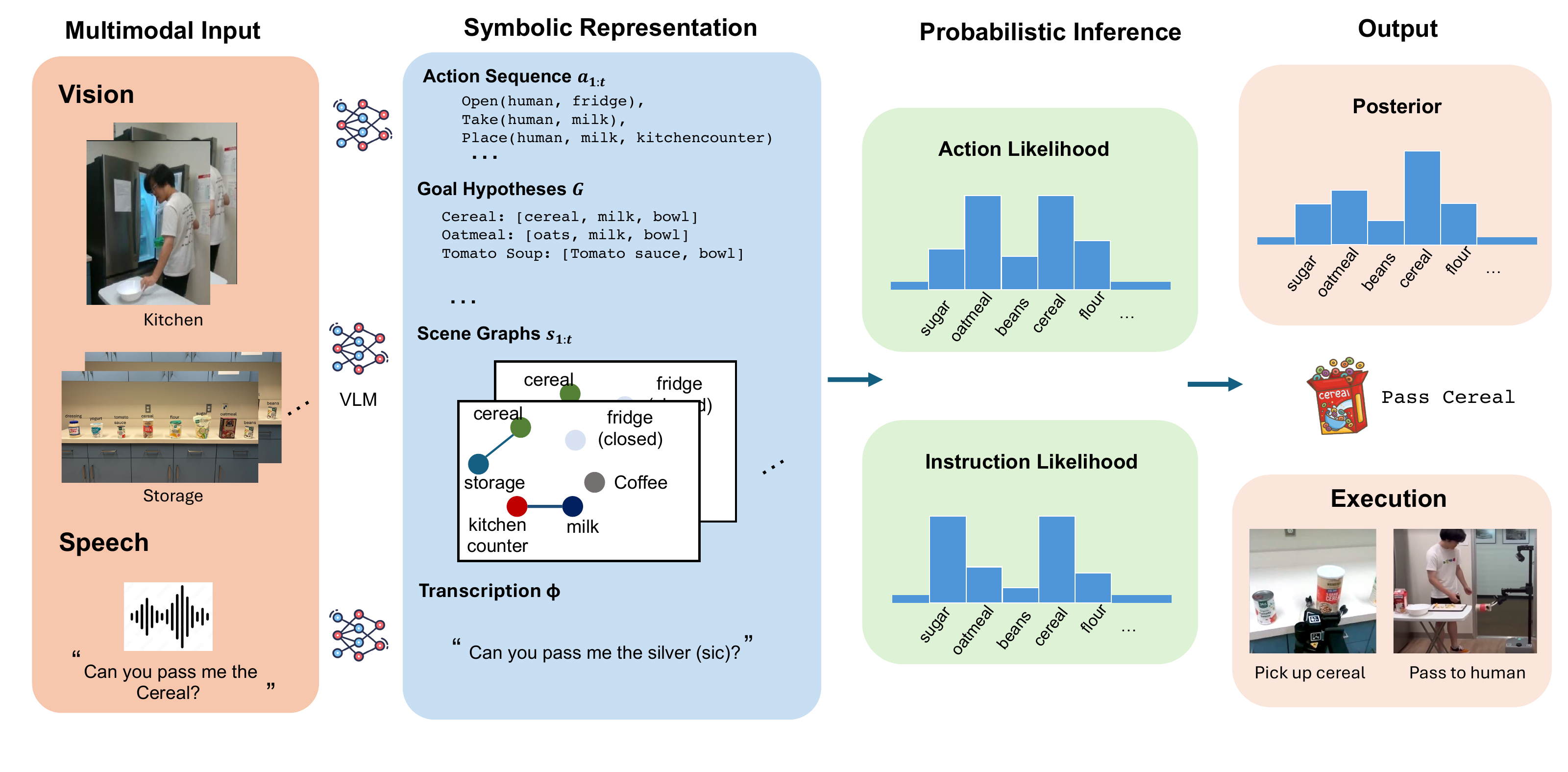}
    \caption{Illustration of the SIFToM architecture for robust instruction following under noise. The system processes raw visual and speech inputs using a Vision-Language Model (VLM) to generate a structured, symbolic representation of the scene, human actions, and transcribed speech. This symbolic data then feeds into a probabilistic inference module that reasons over multiple goal hypotheses to find the most likely human intent. In the example shown, a person retrieves milk and a bowl in the kitchen, then gives a verbal instruction to the robot assistant:  ``Can you pass me the cereal?''. However, due to background noise, the speech is not transcribed accurately by the VLM model, and the model instead outputs ``silver'' instead of ``cereal''. In this example, the model cannot infer the true command based on the visual observation or speech alone: the actions of getting milk and a bowl could suggest the need for cereal or oatmeal. Based on the speech alone, the model finds sugar and cereal to be likely candidates. However, by integrating the two likelihoods, the model is able to infer the true instruction for getting the cereal.}
    \label{fig:siftom}
\end{figure*}




\section{Related Work}

\subsection{Vision Language Model for Human Robot Collaboration}

Recent work shows VLMs can enhance human-robot collaboration by grounding multimodal inputs in context. Many works build modular and cooperative multi-agent systems by leveraging VLMs for planning, communication, and open-ended skill learning \cite{zhang2023building,li2023camel,wang2023voyager}. There have also been dialogic reasoning and centralized assignment pipelines based on VLMs that can improve team performance, adaptability, and human-in-the-loop coordination via verbal communication \cite{mandi2024roco,liu2025coherent}. To generate long-horizon robot plans given human instructions, prior works have developed hierarchical task-to-motion planning methods that leverage VLMs to decompose high-level instructions into structured subgoals and low-level skills \cite{yin2025partinstruct,lin2023text2motion,shi2025hi,wei2025hierarchical}. Furthermore, recent works have introduced socially aware navigation \cite{shah2023lm,song2024vlm,long2024instructnav}, where VLM-based scoring ensures robots act politely and safely in shared spaces. 

\subsection{Language Grounding in Embodied Interactions}

There has been a long history of work attempting to ground unspecified or ambiguous language instructions in embodied environments. Previous work has applied Vision and Large Language Models (LLMs) to map words to real-world objects and subsequent robot actions \cite{min2021film, liu2022instruction, blukis2020few, liu2023lang2ltl, liu2024lang2ltl2, cohen2024ground}. Some recent work in cognitive science and AI has applied computational Theory of Mind models to clarify ambiguous human instructions by reasoning about the human speaker's intent and the joint human-robot plan \cite{zhi2024pragmatic}. However, most language instruction grounding works do not take speech input, and few studies have examined how robust these systems are in noisy environments where instructions may not be fully intelligible.

\subsection{Theory of Mind for Cooperative Robot Planning}

There has been extensive research on inferring other agents' mental states, such as goals, desires, and beliefs (e.g., \cite{zhi2020online, shu2021agent, jin2024mmtom, zhi2024pragmatic}), commonly referred to as Theory of Mind reasoning, to better cooperate with others in collaborative tasks. Earlier work in human-robot interaction has used model-based approaches such as plan recognition and inverse planning to infer other agents' goals. Previous work in cognitive science and psychology has found that Theory of Mind is critical in agents' interactions and cooperation. Computational work in AI and Robotics has also shown that embedding a Theory of Mind module in embodied AI agents often leads to more effective communications and more efficient collaborations between AI and human users \cite{ying2024goma, puig2020watch, puig2023nopa, devin2016implemented}. 

\section{Methods}

\subsection{Problem Formulation}

Following prior work by \cite{zhi2024pragmatic}, we formulate the human-robot collaboration task as an extension of assistance games \cite{hadfield2016cooperative,fisac2019pragmatic} with noisy spoken instructions. We define this as a two-agent Partially observable Markov decision process (POMDP) with a human principal and an assistive agent, described by the tuple $(\mathsf{S}, \mathsf{\Phi}, \mathsf{A}^{_h}, \mathsf{A}^{_r}, \mathsf{C}, \mathsf{G}, H, P_s, P_i)$, where $\mathsf{S}$ is the set of environment states, $\mathsf{\Phi}$ a space of instructions $\phi$ that the human may give to the robot, $\mathsf{A}^{_h}$ the set of human actions $a^{_h}$, $\mathsf{A}^{_r}$ the set of assistant actions $a^{_r}$, $\mathsf{C}$ a set of cost functions $C: \mathsf{S} \times \mathsf{\Phi} \times \mathsf{A}^{_h} \times \mathsf{A}^{_r} \to \mathbb{R}$ that map state-action transitions to real numbers, $\mathsf{G} \subseteq \mathcal{P}(\mathsf{S})$ a set of possible goals $g$ where each $g \subseteq S$ is a set of (terminal) states, $H$ a horizon after which the game automatically terminates, $P_s(s'|s, a^{_r}, a^{_h})$ the environment transition distribution, and $P_i(s_i, C, g)$ a distribution over initial states of the game. In our assistance games, the human knows the \emph{true} cost function $C$ and goal $g$, as well as the instruction given to the robot, but the assistant can only observe the environment state $s_t$. Thus, the assistant has to \emph{infer} the true goal $g$ and the instruction $\phi$, which may be corrupted by noise.

\subsection{Spoken Instruction Following through Theory of Mind (SIFToM)}

\subsubsection{Model Overview}

The neurosymbolic architecture of the SIFToM model is shown in Figure \ref{fig:siftom}. The model has two key components. 1) SIFToM first prompts the VLM to parse visual and speech inputs into a symbolic representation. 2) The model then performs probabilistic Theory of Mind inference conditioned on the symbolic representations to find the most likely goal and command intended by the human user. We describe the two components in detail below.

\subsubsection{Generating Symbolic Representations}
In order to perform model-based inference over the agent, we first need to generate a model of the agent and the task. SIFToM follows prior work by \cite{jin2024mmtom, ying2025language}, which prompts a VLM to generate symbolic representations for the task, including the following components:

\begin{itemize}
    \item \textbf{Scene Graphs:} Scene graphs, denoted as $s_t$, are structured representations of the environment's state at a given timestep $t$. Each graph captures key objects, their properties (e.g., \texttt{closed(fridge)}), and their spatial or semantic relationships (e.g., \texttt{on(milk, kitchencounter)}). The VLM parses visual input from the scene to generate and update these graphs over time, forming a state trajectory $s_{1:t}$.

    \item \textbf{Goal Space:} The goal space $G$ is a discrete set of plausible high-level tasks the human may be pursuing, generated by the VLM based on the overall environmental context. For instance, in a kitchen setting, the VLM might hypothesize goals such as \texttt{make\_cereal} or \texttt{make\_french\_toast}. Each goal $g \in G$ can be defined by a set of required objects or target states, and this set serves as the prior hypothesis space for our downstream inference module.

    \item \textbf{Action Sequence:} The action sequence $a_{1:t}$ is a time-ordered, symbolic log of the human's physical interactions with objects in the environment. The VLM observes the human's movements from video frames and discretizes them into a structured \texttt{Predicate(Agent, Object, Location)} format, for example, place(human, milk, kitchencounter) represents the human action of placing the milk on the kitchencounter.

    \item \textbf{Instruction:} The instruction, denoted by $\phi$, is the textual transcription of the human's verbal command as interpreted by the VLM's automatic speech recognition (ASR) module. We explicitly treat this component as a potentially noisy signal. As illustrated in Figure 1, environmental factors or ASR limitations can lead to transcription errors (e.g., ``cereal'' being misheard as ``silver''), making the raw instruction an unreliable signal when considered in isolation.
\end{itemize}

\subsubsection{Probabilistic Goal and Plan Inference}

Once we have a symbolic environment model, SIFToM can leverage both visual observations and the transcribed (potentially corrupted) instruction to jointly infer the team's goal $g$ and the intended joint plan $\pi = (\pi_h, \pi_r)$, where $\pi_h$ refers to the human plan and $\pi_r$ is the robot's plan. The full posterior over goals and plans given all observations can be expressed using Bayes' rule:
\begin{align} 
    P(&g, \pi|s_{1:T}, \phi, a^{h}_{1:T}, a^{r}_{1:T}) \nonumber \\ 
    &\propto P(g, \pi, s_{1:T}, \phi, a^{h}_{1:T}, a^{r}_{1:T}) \nonumber 
    \\ 
    &= P(g, \pi) \textstyle{\prod}_{t=1}^{T}  P(s_t | s_{t-1}, a^{h,r}_{t-1}) P(\phi| s_t, \pi) P(a^{h}_t, a^{r}_t|s_t, \pi). \label{eq:full_posterior}
\end{align}
In this formulation, $P(g, \pi)$ is the prior over goals and plans, and $P(s_t | s_{t-1}, a^{h,r}_{t-1})$ is the state transition model, which is deterministic in our symbolic representation. The inference problem thus centers on computing the two key likelihood terms: the action likelihood $P(a^{h}_t, a^{r}_t|s_t, \pi)$ and the instruction likelihood $P(\phi| s_t, \pi)$. Our objective is to find the goal $g^*$ and the corresponding optimal plan $\pi^*$ that maximizes this posterior.

SIFToM is a general architecture that enables VLMs to perform model-based probabilistic social inference in collaborative settings, and there can be many ways to compute or estimate these two likelihood functions inside the probabilistic framework. The novelty of the model lies in the modular approach, integrating uncertainty from different sources within an explicit neurosymbolic probabilistic framework. In the sections below, we describe one concrete implementation for the likelihood functions for our set of experiments. 

\paragraph{Action Likelihood}
The term $P(a^{h}_t, a^{r}_t|s_t, \pi)$ represents the likelihood that the agents execute the actions $a^h_t$ and $a^r_t$ in state $s_t$, given that they are following the joint plan $\pi$. We are primarily concerned with the human's actions $a^h_t$ as the source of evidence, as the robot has stayed idle before receiving the spoken instruction. We model the human's adherence to their part of the plan, $\pi_h$, using a Boltzmann rationality model:
\begin{align}
    P(a^h_t|s_t, \pi) \propto \exp(\beta \cdot Q^*(s_t, a^h_t; \pi_h)),
    \label{eq:action_likelihood}
\end{align}
where $\beta$ is a rationality parameter and $Q^*(s_t, a^h_t; \pi_h)$ is the value of taking action $a^h_t$ in state $s_t$ while following plan $\pi_h$. This quantifies how likely the observed human action is if they are indeed executing the hypothesized plan. To estimate $Q^*(s_t, a^h_t; \pi_h)$, we adopt prior work in online goal inference \cite{zhi2020online}, which uses A-star planning to compute policies on-the-fly.

\paragraph{Instruction Likelihood}
The term $P(\phi| s_t, \pi)$ models the probability of observing the utterance $\phi$ given the current state and the joint plan. We posit that the human's instruction is primarily directed at the robot, serving as a communication of the expected actions to take in the robot's plan, $\pi_r$. The instruction $\phi$ is therefore a (potentially noisy) label for the robot's plan. The likelihood is thus conditioned on the robot's component of the joint plan:
\begin{align}
    P(\phi| s_t, \pi) = P(\phi| s_t, \pi_r).
\end{align}
We compute this probability using the token log probabilities from a VLM, which scores the relevance of the utterance $\phi$ to a textual description of the robot's planned actions $\pi_r$, when prompted with a few examples of corrupted instructions and their corresponding ground-truth commands. For example, with few-shot prompting, the VLM evaluates how likely the transcribed utterance ``pass me the silver'' corresponds to the command of executing \texttt{pass(robot, human, cereal)}. This allows the model to ground the ambiguous language in the concrete action space of the robot, iterating through all possible objects in the environment and finding the command that best matches the noisy instruction.

By computing these likelihoods at each step and integrating them as shown in Eq.~\ref{eq:full_posterior}, SIFToM can perform a robust inference over the intended goal and plan, even in the presence of ambiguous actions and corrupted linguistic input.

\section{Simulated Experiment}

As we were not able to find any existing noisy spoken instruction dataset for embodied collaboration, we first constructed a novel noisy speech instruction dataset, UnclearInstruct, in a simulated home environment where a human principal agent requests help from an AI agent to complete a household task. This simulated experiment allows us to systematically stress test a model's robustness under different kinds of noise and tasks in a controlled setting.

\subsection{Dataset Construction}

The UnclearInstruct dataset is situated in a simulator household environment with real human speech instructions. 

\begin{figure}
    \centering
    \includegraphics[width=0.9\linewidth]{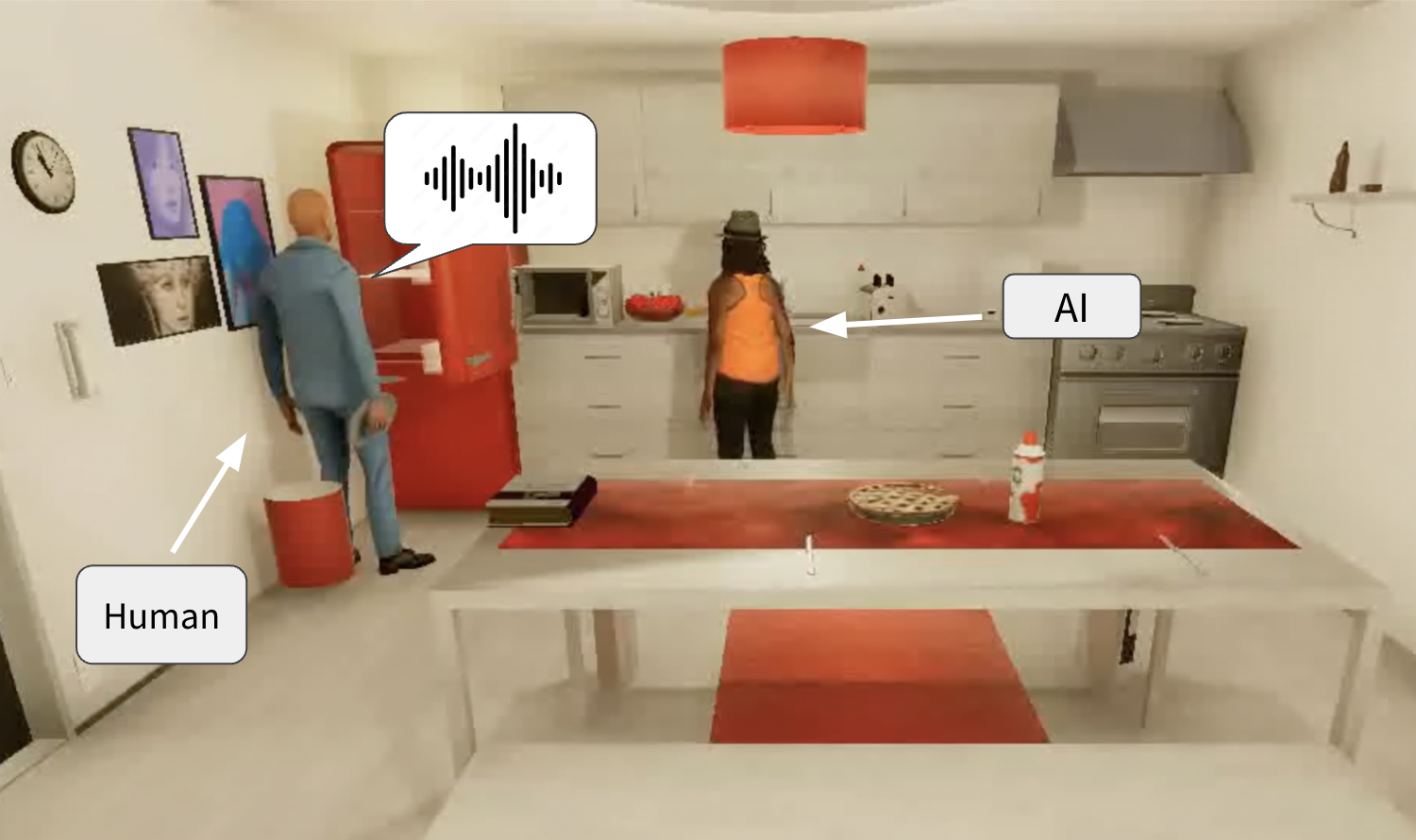}
    \caption{An illustration of the VirtualHome simulator where a main agent and an assistant collaborate on a household task.}
    \label{fig:vh-img}
    \vspace{-5pt}
\end{figure}

\subsubsection{Tasks and Scenarios}

We sampled 26 task scenarios from VirtualHome 2~\cite{puig2020watch}, a multiagent household embodied simulator (Figure~\ref{fig:vh-img}). In this environment, the human and robot agent can pick up or place objects, open containers, and move around the room. The task scenarios in UnclearInstruct feature a human agent and a robot assistant collaborating to set up tables in a dining room. The human agent is simulated by an MCTS planner. The human agent's goals and the goal space are hidden from the robot assistant.

The environments include typical household items such as plates, forks, knives, wineglasses, etc. The tasks involve picking up and placing different objects with the same object count(e.g., placing 3 forks, 3 knives, and 3 plates on the table), but the object count, the types of objects needed, the location of the objects, and the room setup can vary across scenarios.

\subsubsection{Human Instruction Collection}

We recruited 5 participants (Mean age: 24.5, 3 Male, 2 Female) from a US university. All participants are non-native English speakers. We showed the participants each task scenario and asked them to provide natural-language instructions they would give to the robot in each scenario. We then recorded participants speaking these instructions. In total, we obtained 130 speech instructions. 

\subsubsection{Noisy Speech Instruction Generation}

To simulate real-world speech instructions under noisy conditions, we follow prior ASR work by injecting noise into the original speech by merging real-world noise audio with human speech instructions. We used the noise files from the Microsoft Scalable Noisy Speech Dataset (MS-SNSD) dataset \cite{reddy2019scalable}, which included 14 types of real-world noise clips, such as air conditioner, appliance noise, car noise, multi-talker babble, etc. 

The final dataset has 520 stimuli, including 130 stimuli with the original human speech, and 390 stimuli with noise-augmented speech instructions.

\subsection{Models and Baselines}

\subsubsection{Human baseline}
To evaluate the human performance on the dataset, we ran a human annotation experiment with an online human interface. We recruited 130 human participants (mean age = 35.08; 84 female, 46 male) over Prolific to annotate the stimuli. In each trial, the participant was shown the scenario and assumed the role of the assistant (recipient of the speech instruction). After viewing the scenario, the human participants were asked to give a transcription of the spoken instruction and select the items they would pick up to aid the main agent. This experiment was approved by an institutional review board.

\subsubsection{SIFToM and Ablation}
We instantiate the SIFToM model described in the previous section using Gemini 2.5 Flash, which is a relatively lightweight, fast Vision-Language Model.

We also consider an ablated SIFToM model, where we feed the same base Gemini 2.5 Flash model with the symbolic representation parsed from multimodal inputs. However, instead of performing explicit model-based probabilistic Theory of Mind inference, the model is instead prompted to reason about the human intent while transcribing the speech instruction and producing the output.

\subsubsection{VLM baselines}
We consider two end-to-end baselines: Gemini 2.5 Flash and Gemini 2.5 Pro. We use the Gemini 2.5 Flash model as it is the same base model used in our SIFToM implementation. This allows us to evaluate whether the neurosymbolic framework introduced in SIFToM can leverage a lightweight neural model to solve challenging instruction following tasks. Additionally, we use Gemini 2.5 Pro, a state-of-the-art VLM, as a strong neural baseline.

\subsection{Performance Metrics}

We evaluated the models with two main performance metrics: accuracy rate and speedup. We also include model runtime and token counts in the analysis.

\paragraph{Accuracy} We measured inference accuracy by the percentage of trials in which the robot inferred the correct robot goal from speech instructions.

\paragraph{Speedup}
Following past work on human-robot collaboration in VirtualHome \cite{puig2023nopa}, we computed the speedup of the baseline models against a single-agent (human alone) baseline. The speedup is computed as $ Speedup = L_{single}/L_{team} - 1$, where $L_{single}$ is the timesteps it takes for the human agent to complete the task alone without robot assistance and $L_{team}$ is the timesteps for the human-robot team. In cases where the robot assistant misunderstands the command and executes the incorrect actions, we simulate and include any additional steps the humans need to take to remedy the situation, including removing the incorrect items and replacing them with the correct ones, which could increase the total timesteps it takes for the team to complete the task.


\subsection{Results}

\begin{figure}
    \centering
    \includegraphics[width=1\linewidth]{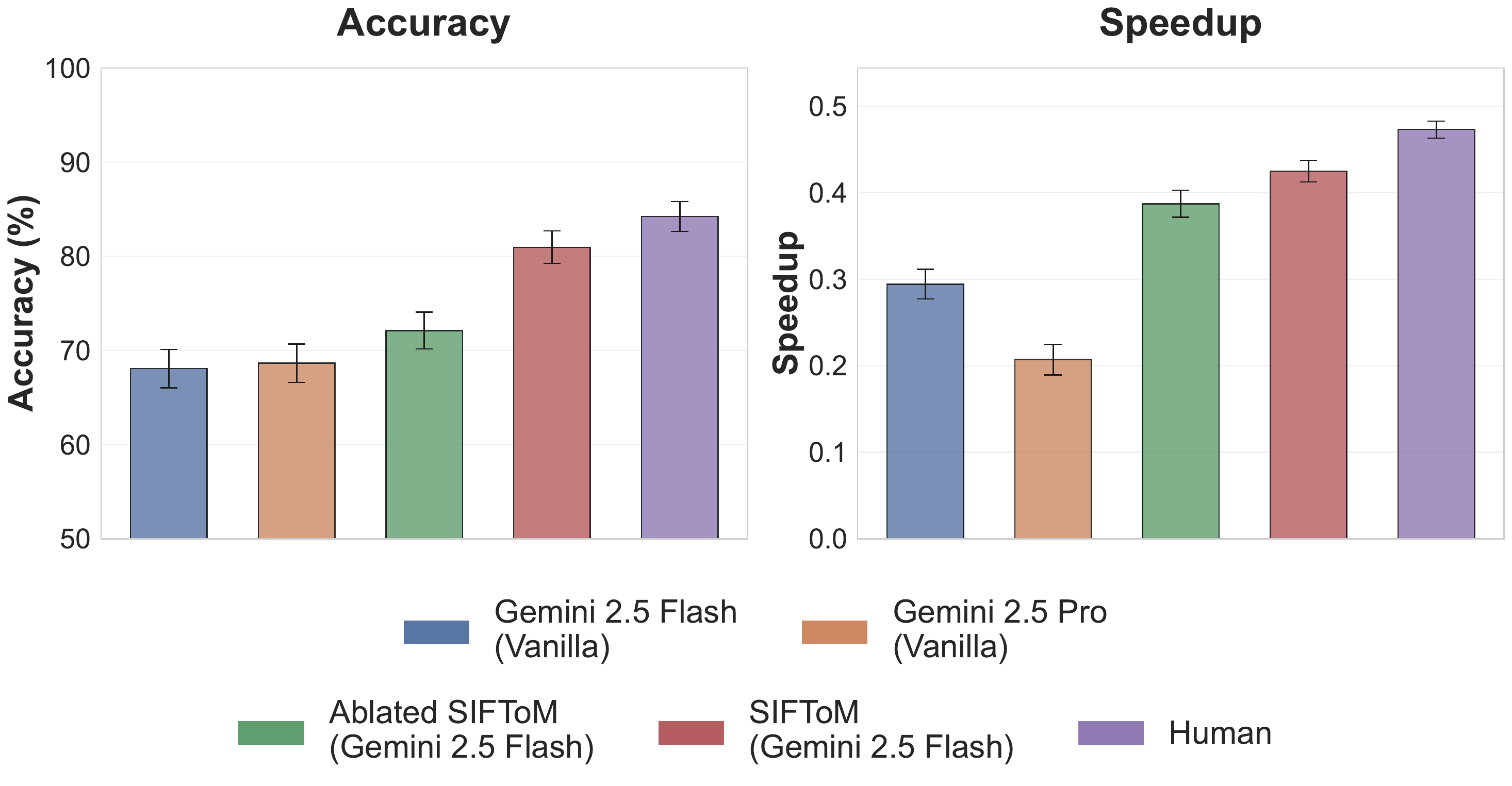}
    \caption{Model accuracy and speedup performance. Overall, SIFToM achieved the best performance among all models and approached human performance. Error bars indicate 95\% confidence interval bootstrapped from 1000 samples.}
    \label{fig:sim-exp}
    \vspace{-5pt}
\end{figure}

\begin{figure}
    \centering
    \includegraphics[width=1\linewidth]{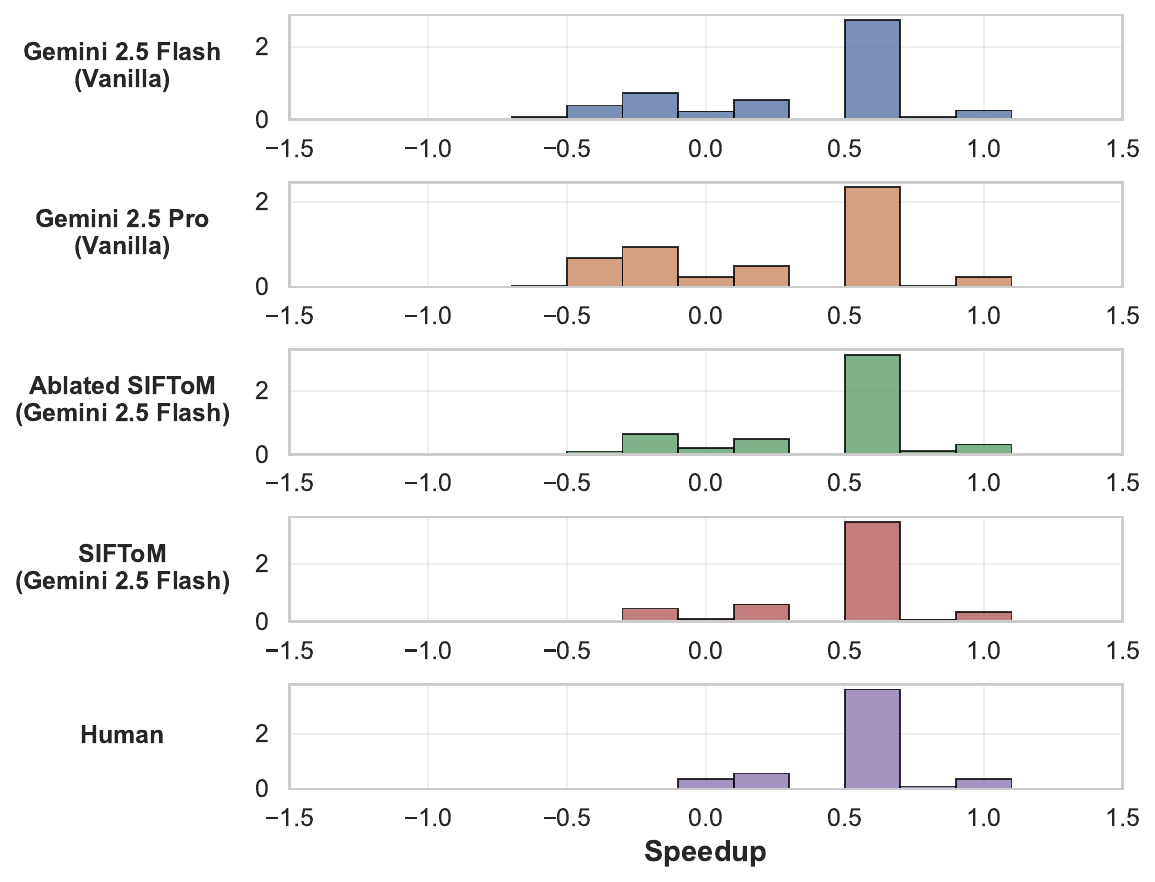}
    \caption{Histograms showing the distribution of speedup metrics across models.}
    \label{fig:speedup_dist}
    \vspace{-5pt}
\end{figure}

We report the model performances in Figure~\ref{fig:sim-exp}. Overall, we found that Gemini 2.5 Pro accuracy (mean = 68.7\%) was similar to Gemini 2.5 Flash (mean = 68.1\%). However, SIFToM had a 12.9\% increase in accuracy and 44.5\% increase in speedup over the base model, significantly outperforming Gemini 2.5 Pro and similar to human performance level. We noticed a similar trend for speedup. The difference between Gemini 2.5 Flash and Pro vanilla was surprising, as the Pro model performed worse on the speedup metrics, even though the accuracy scores were similar.


\begin{figure*}[ht!]
    \centering
    \includegraphics[width=1\linewidth]{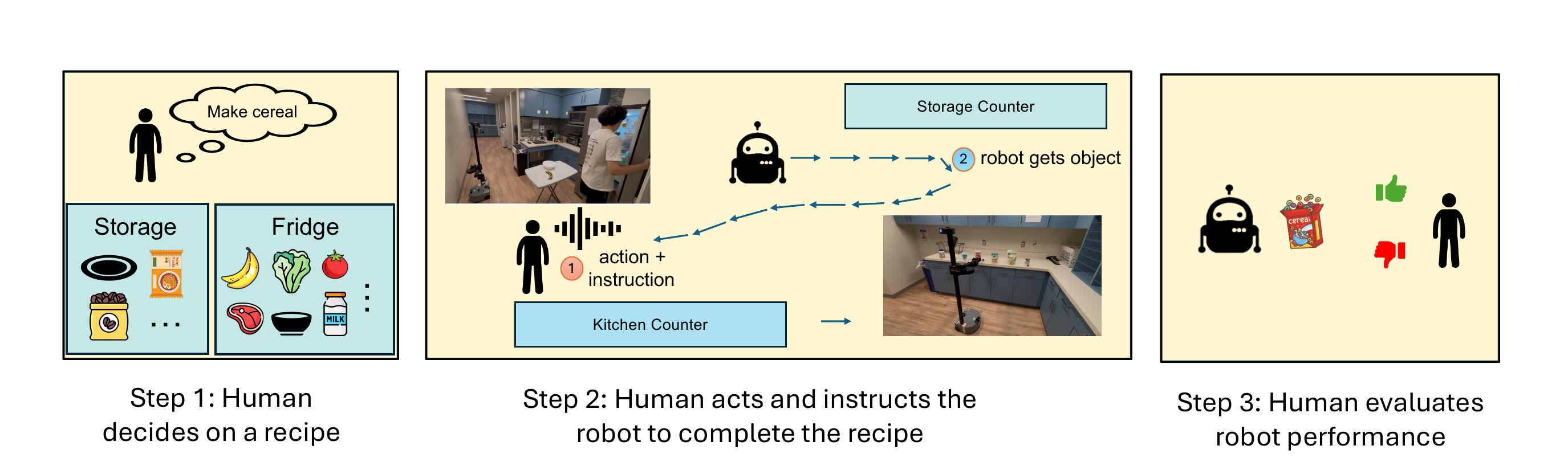}
    \caption{Illustration of the experimental setup for the real-world robot experiment. Each trial of the human-robot experiment follows three steps. First, a human subject is asked to come up with a goal recipe when given a set of ingredients in the environment. Then the human participant starts preparing for the recipe while instructing the robot to help collect some additional food items with some noise in the background. Lastly, when the trial ends, the human participant rates the performance of the robot.}
    \label{fig:human_exp_setup}
\end{figure*}

\paragraph*{\textbf{Error Analysis}} We performed an error analysis and found that the mistakes made by humans or SIFToM were less costly than those made by other models. Upon analyzing the incorrect guesses, we found that 100\% of incorrect human guesses were, in fact, valid and helpful actions, meaning that the items inferred were part of the team goal, but often the item count was incorrect. This ratio dropped to 69.7\% for Gemini 2.5 Flash with SIFToM and less than 40\% for all other models. 

Figure~\ref{fig:speedup_dist} shows the distribution of speedups. We found that a significant portion of trials in the Gemini 2.5 Flash (24.2\%) and Pro models (34.2\%) have a negative speedup, which indicates that the robot actions severely hindered the team performance. In contrast, such trials are rare in humans (0.4\%) and SIFToM (11.2\%). This shows that, similar to humans, by grounding speech instructions in the inferred joint goal, SIFToM is more likely to perform useful actions even when it doesn't have the exact command inferred.

\section{Real-World Robot Experiment}

To evaluate whether SIFToM can enable a robot to assist humans in the real world noisy environments, we conducted a real-world robot experiment where a human participant worked with a Stretch robot to prepare a meal together. We recruited 15 participants from a US university (Mean age = 23.2; 11 male, 4 female) for the experiment. The experiment was approved by an institutional review board.

\subsection{Experimental Setup}

As shown in Figure~\ref{fig:human_exp_setup}, we conducted the real-world robot experiment in a mock home environment, where participants were instructed to prepare a meal with a Stretch robot. The environment has a kitchen space and a storage space. Some items are located in the kitchen, and some are on the storage counter. There are 39 food items in total. 

The participants were first shown all food items available in the environment. The participants were then asked to come up with a recipe they wanted to make and coordinate with the robot to prepare the meal through language instruction, and they were not given any recipes to choose from. The ground truth goal recipes that participants came up with were unknown to the robot. Note that participants were never instructed or prompted to come up with specific goals. This setting ensures the open-endedness and the realism of the task as well as the user behavior in the experiment. Table~\ref{tab:embodied-goal} shows example goal recipes provided by human participants. In total, 109 unique goal tasks were generated by human participants out of 150 trials. 

There were three experimental conditions, corresponding to the three different models embedded into the robot assistant. The experiment adopted a within-subject design, where human participants completed ten trials with a mix of experimental conditions in a randomized order. To balance among the experimental conditions, we made sure that each human participant encountered each model at least twice in the ten trials. The human participants were not aware which model was used in each trial. To simulate a noisy collaborative environment, we injected background noise in the experiment through an external speaker.

At the end of each trial, the human participants were asked to report their goal recipe and the instructions they gave to the robot. They then indicated whether the robot had performed the task as expected, as well as their subjective evaluation of the robot's performance on the task.

\begin{table}[t]
    \centering
\begin{tabular}{P{2cm} P{6cm}}
\toprule
\textbf{Sample Goals} & \textbf{Goal Specification} \\
\midrule
Breakfast Cereal & Cereal, Bowl, Milk\\
Prepare Coffee & Coffee, Sugar, Cup, Milk\\
Prepare Tea & Tea, Cup, Milk\\
Tomato Pasta & Pasta, Tomato Sauce, Bowl\\
Vegetable Salad & Lettuce, Cucumber, Dressing, Plate \\
Prepare Pancake & Egg, Flour, Milk, Bowl\\
Bacon Sandwich & Dressing, Bread, Lettuce, Bacon\\
\bottomrule

\end{tabular}
    \caption{We showcase 7 example goal recipes out of 109 unique recipes proposed by human participants among 150 trials of a real-world robot experiment based on the available ingredients in the environment.}
    \label{tab:embodied-goal}
    \vspace{-5pt}
\end{table}

\subsection{Robot Command Execution}
For the robot to execute the command inferred by each model, the Stretch robot first generates a motion plan towards the target object using an A-star planner \cite{hart1968formal}, then the Stretch robot uses AnyGrasp \cite{fang2023anygrasp} to grasp the target object and deliver it to the human.

\subsection{Models and Baselines}

As the evaluation of each model in the real-world experiment requires a significant amount of human subject trials, we only included SIFToM (with Gemini 2.5 Flash), ablated SIFToM (Gemini 2.5 Flash), and Gemini 2.5 Pro.

\subsection{Performance Metrics}

We use the following two metrics for evaluating the models' performance.

\begin{itemize}
    \item Success rate: Success rate measures if the inferred robot task matches the one intended by the human participant in each trial.
    \item Human subjective rating: In addition to the objective correctness of the robot execution, we also ask human participants to rate three statements on a 7-point Likert Scale (1 = Strongly Disagree, 7 = Strongly Agree): 1) ``The robot performed the task as expected, considering the difficulty of the task.'' 2) ``The robot understood what I needed.'' 3) ``The robot performed on par with or better than what I would expect from a human.'' We take the average of these three values as a measure of model helpfulness. We include these subjective evaluations as the trials could vary in difficulty. The human participants' evaluations are designed to take the task difficulty into account when evaluating model performance, beyond just raw success rate.  
\end{itemize}

\subsection{Results}

\begin{figure}
    \centering
    \includegraphics[width=1\linewidth]{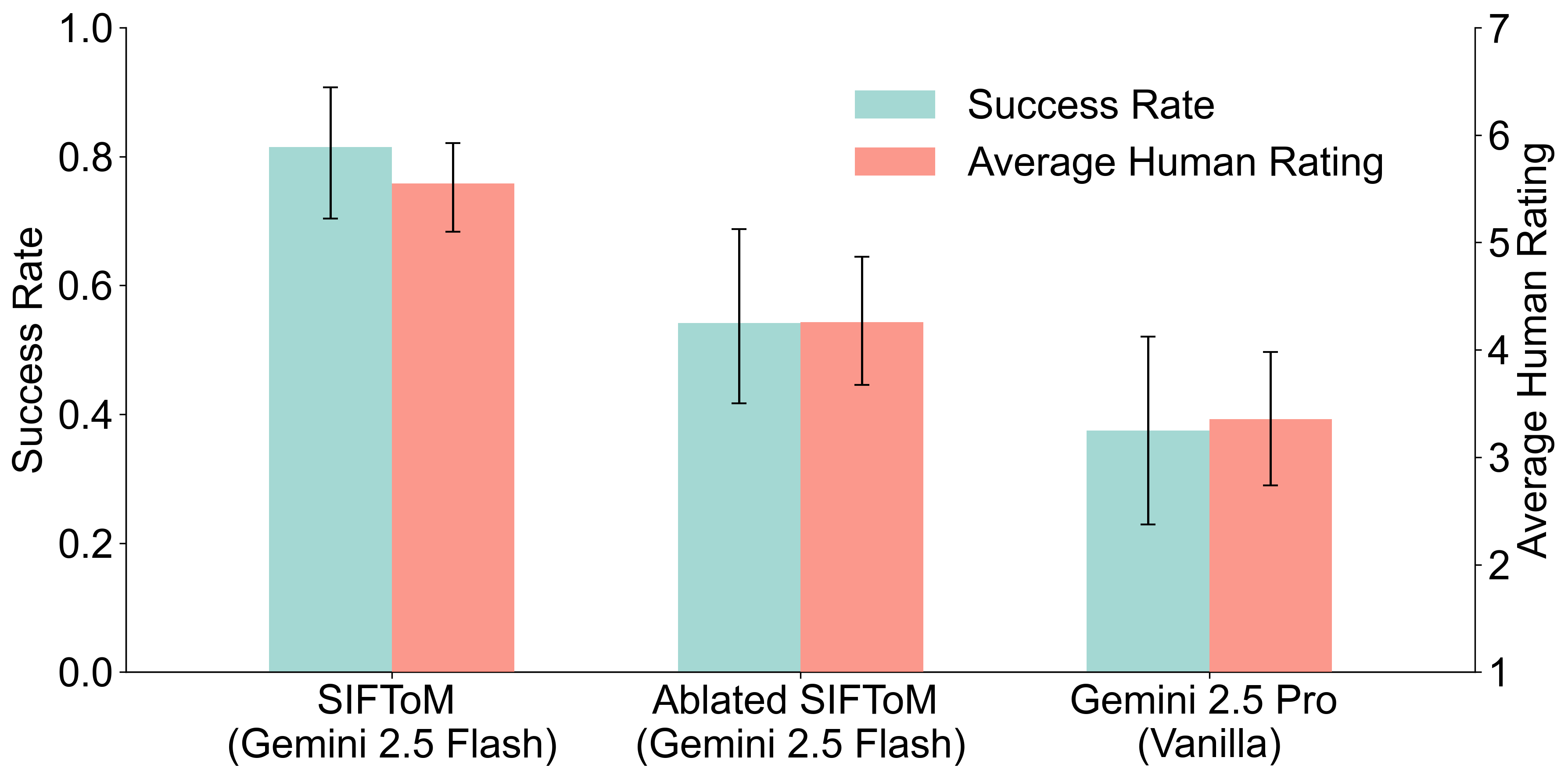}
    \caption{Success rate and human subjective ratings in the Human-robot experiment. Error bars indicate 95\% confidence interval bootstrapped from 1000 samples.}
    \label{fig:real-exp}
    \vspace{-5pt}
\end{figure}
The results of the real-world robot experiment are shown in Figure \ref{fig:real-exp}. Overall, the SIFToM model with Gemini Flash 2.5 was able to complete 81.48\% of the trials correctly (SD = 0.39), significantly higher than the ablated SIFToM version (mean = 54.17\%, SD = 0.50, p = \(2.72\times 10^{-3}\)), and Gemini 2.5 Pro (mean = 37.50\%, SD = 0.48, p =\(2.13\times 10^{-6}\)). SIFToM performance was comparable to the simulated experiment, whereas Gemini 2.5 Pro performance degraded significantly. This might be due to the increased complexity in the real-world scenario with more objects and a significantly larger goal space. Similarly, we find that human participants rated SIFToM performance much higher (mean = 5.55, SD = 1.54) than the ablated model (mean = 4.26, SD = 2.23) and Gemini 2.5 Pro (mean = 3.35, SD = 2.19). We provide an example in the supplementary video.



\section{Discussion and Conclusion}

In this paper, we introduced SIFToM, a cognitively-inspired neurosymbolic architecture designed to bridge the gap between the capabilities of modern Vision-Language Models (VLMs) and the demands of real-world human-robot collaboration. SIFToM enables robots to ground and disambiguate noisy human instructions by reasoning about intent through a Theory of Mind framework. Rather than relying on literal interpretations of potentially corrupted speech, our model integrates multimodal context, using the human's physical actions to infer the true goal behind their words.

Our experiments in both simulated and embodied environments demonstrate that this approach is critical for robust interaction. SIFToM consistently outperformed VLM baselines and ablated versions in correctly inferring human intent from ambiguous inputs. Furthermore, our error analysis reveals that when SIFToM does fail, it does so more gracefully, making contextually plausible, less costly mistakes, in contrast to the random guesses of other models. This pragmatic inference capability is essential for deploying robots that can be trusted to act safely and predictably alongside people.

However, our work also highlights a key bottleneck for developing robust robot partners for human users in the wild: the fidelity of their perceptual and symbolic grounding. The performance gap between SIFToM and human partners underscores that even with a robust inference engine, the agent's effectiveness is limited by the quality of its underlying visual parser. Errors in translating the rich, continuous real world into a discrete, symbolic representation remain a primary source of failure.

Future work should therefore focus on improving the robustness of this grounding process and on extending the SIFToM framework to more challenging, ``in-the-wild'' settings. This includes testing the model in environments with greater visual and acoustic complexity, such as factories or crowded public spaces, and with a wider diversity of human partners, including those with strong accents or speech disorders.

We believe SIFToM offers a promising path toward VLM-powered agents that can move beyond controlled settings and become truly effective collaborators. By equipping them with the ability to pragmatically reason about human intent in context, we can build robots that are not only more intelligent but also more resilient and better aligned with the fluid, imperfect nature of human interaction in the real world.

\bibliography{custom}

\begin{thebibliography}{10}
\providecommand{\url}[1]{#1}
\csname url@samestyle\endcsname
\providecommand{\newblock}{\relax}
\providecommand{\bibinfo}[2]{#2}
\providecommand{\BIBentrySTDinterwordspacing}{\spaceskip=0pt\relax}
\providecommand{\BIBentryALTinterwordstretchfactor}{4}
\providecommand{\BIBentryALTinterwordspacing}{\spaceskip=\fontdimen2\font plus
\BIBentryALTinterwordstretchfactor\fontdimen3\font minus \fontdimen4\font\relax}
\providecommand{\BIBforeignlanguage}[2]{{%
\expandafter\ifx\csname l@#1\endcsname\relax
\typeout{** WARNING: IEEEtran.bst: No hyphenation pattern has been}%
\typeout{** loaded for the language `#1'. Using the pattern for}%
\typeout{** the default language instead.}%
\else
\language=\csname l@#1\endcsname
\fi
#2}}
\providecommand{\BIBdecl}{\relax}
\BIBdecl

\bibitem{tomasello2010origins}
M.~Tomasello, \emph{Origins of human communication}.\hskip 1em plus 0.5em minus 0.4em\relax MIT press, 2010.

\bibitem{ahn2022can}
B.~Ichter, A.~Brohan, Y.~Chebotar, C.~Finn, K.~Hausman, .~..., and C.~Kelly, ``Do as i can, not as i say: Grounding language in robotic affordances,'' in \emph{Proceedings of The 6th Conference on Robot Learning}, ser. Proceedings of Machine Learning Research, K.~Liu, D.~Kulic, and J.~Ichnowski, Eds., vol. 205.\hskip 1em plus 0.5em minus 0.4em\relax PMLR, 14--18 Dec 2023, pp. 287--318.

\bibitem{gao2024physically}
J.~Gao, B.~Sarkar, F.~Xia, T.~Xiao, J.~Wu, B.~Ichter, A.~Majumdar, and D.~Sadigh, ``Physically grounded vision-language models for robotic manipulation,'' in \emph{2024 IEEE International Conference on Robotics and Automation (ICRA)}.\hskip 1em plus 0.5em minus 0.4em\relax IEEE, 2024, pp. 12\,462--12\,469.

\bibitem{puig2020watch}
X.~Puig, T.~Shu, S.~Li, Z.~Wang, Y.-H. Liao, J.~B. Tenenbaum, S.~Fidler, and A.~Torralba, ``Watch-and-help: A challenge for social perception and human-ai collaboration,'' \emph{arXiv preprint arXiv:2010.09890}, 2020.

\bibitem{zhang2023building}
H.~Zhang, W.~Du, J.~Shan, Q.~Zhou, Y.~Du, J.~B. Tenenbaum, T.~Shu, and C.~Gan, ``Building cooperative embodied agents modularly with large language models,'' \emph{arXiv preprint arXiv:2307.02485}, 2023.

\bibitem{li2023camel}
G.~Li, H.~Hammoud, H.~Itani, D.~Khizbullin, and B.~Ghanem, ``Camel: Communicative agents for" mind" exploration of large language model society,'' \emph{Advances in Neural Information Processing Systems}, vol.~36, pp. 51\,991--52\,008, 2023.

\bibitem{wang2023voyager}
G.~Wang, Y.~Xie, Y.~Jiang, A.~Mandlekar, C.~Xiao, Y.~Zhu, L.~Fan, and A.~Anandkumar, ``Voyager: An open-ended embodied agent with large language models,'' \emph{arXiv preprint arXiv:2305.16291}, 2023.

\bibitem{mandi2024roco}
Z.~Mandi, S.~Jain, and S.~Song, ``Roco: Dialectic multi-robot collaboration with large language models,'' in \emph{2024 IEEE International Conference on Robotics and Automation (ICRA)}.\hskip 1em plus 0.5em minus 0.4em\relax IEEE, 2024, pp. 286--299.

\bibitem{liu2025coherent}
K.~Liu, Z.~Tang, D.~Wang, Z.~Wang, X.~Li, and B.~Zhao, ``Coherent: Collaboration of heterogeneous multi-robot system with large language models,'' in \emph{2025 IEEE International Conference on Robotics and Automation (ICRA)}.\hskip 1em plus 0.5em minus 0.4em\relax IEEE, 2025, pp. 10\,208--10\,214.

\bibitem{yin2025partinstruct}
Y.~Yin, Z.~Han, S.~Aarya, J.~Wang, S.~Xu, J.~Peng, A.~Wang, A.~Yuille, and T.~Shu, ``Partinstruct: Part-level instruction following for fine-grained robot manipulation,'' \emph{arXiv preprint arXiv:2505.21652}, 2025.

\bibitem{lin2023text2motion}
K.~Lin, C.~Agia, T.~Migimatsu, M.~Pavone, and J.~Bohg, ``Text2motion: From natural language instructions to feasible plans,'' \emph{Autonomous Robots}, vol.~47, no.~8, pp. 1345--1365, 2023.

\bibitem{shi2025hi}
L.~X. Shi, B.~Ichter, M.~Equi, L.~Ke, K.~Pertsch, Q.~Vuong, J.~Tanner, A.~Walling, H.~Wang, N.~Fusai \emph{et~al.}, ``Hi robot: Open-ended instruction following with hierarchical vision-language-action models,'' \emph{arXiv preprint arXiv:2502.19417}, 2025.

\bibitem{wei2025hierarchical}
Z.~Wei, X.~Luo, and C.~Liu, ``Hierarchical temporal logic task and motion planning for multi-robot systems,'' \emph{arXiv preprint arXiv:2504.18899}, 2025.

\bibitem{shah2023lm}
D.~Shah, B.~Osi{\'n}ski, S.~Levine \emph{et~al.}, ``Lm-nav: Robotic navigation with large pre-trained models of language, vision, and action,'' in \emph{Conference on robot learning}.\hskip 1em plus 0.5em minus 0.4em\relax PMLR, 2023, pp. 492--504.

\bibitem{song2024vlm}
D.~Song, J.~Liang, A.~Payandeh, A.~H. Raj, X.~Xiao, and D.~Manocha, ``Vlm-social-nav: Socially aware robot navigation through scoring using vision-language models,'' \emph{IEEE Robotics and Automation Letters}, 2024.

\bibitem{long2024instructnav}
Y.~Long, W.~Cai, H.~Wang, G.~Zhan, and H.~Dong, ``Instructnav: Zero-shot system for generic instruction navigation in unexplored environment,'' \emph{arXiv preprint arXiv:2406.04882}, 2024.

\bibitem{min2021film}
S.~Y. Min, D.~S. Chaplot, P.~Ravikumar, Y.~Bisk, and R.~Salakhutdinov, ``Film: Following instructions in language with modular methods,'' \emph{arXiv preprint arXiv:2110.07342}, 2021.

\bibitem{liu2022instruction}
H.~Liu, L.~Lee, K.~Lee, and P.~Abbeel, ``Instruction-following agents with jointly pre-trained vision-language models,'' 2022.

\bibitem{blukis2020few}
V.~Blukis, R.~A. Knepper, and Y.~Artzi, ``Few-shot object grounding and mapping for natural language robot instruction following,'' \emph{arXiv preprint arXiv:2011.07384}, 2020.

\bibitem{liu2023lang2ltl}
\BIBentryALTinterwordspacing
J.~X. Liu, Z.~Yang, I.~Idrees, S.~Liang, B.~Schornstein, S.~Tellex, and A.~Shah, ``Grounding complex natural language commands for temporal tasks in unseen environments,'' in \emph{Conference on Robot Learning (CoRL)}, 2023. [Online]. Available: \url{https://arxiv.org/abs/2302.11649}
\BIBentrySTDinterwordspacing

\bibitem{liu2024lang2ltl2}
J.~X. Liu, A.~Shah, G.~Konidaris, S.~Tellex, and D.~Paulius, ``{Lang2LTL}-2: Grounding spatiotemporal navigation commands using large language and vision-language models,'' in \emph{IEEE/RSJ International Conference on Intelligent Robots and Systems (IROS)}, 2024.

\bibitem{cohen2024ground}
V.~Cohen, J.~X. Liu, R.~Mooney, S.~Tellex, and D.~Watkins, ``A survey of robotic language grounding: Tradeoffs between symbols and embeddings,'' in \emph{International Joint Conference on Artificial Intelligence}, 2024.

\bibitem{zhi2024pragmatic}
T.~Zhi-Xuan, L.~Ying, V.~Mansinghka, and J.~B. Tenenbaum, ``Pragmatic instruction following and goal assistance via cooperative language-guided inverse planning,'' \emph{arXiv preprint arXiv:2402.17930}, 2024.

\bibitem{zhi2020online}
T.~Zhi-Xuan, J.~Mann, T.~Silver, J.~Tenenbaum, and V.~Mansinghka, ``Online bayesian goal inference for boundedly rational planning agents,'' \emph{Advances in Neural Information Processing Systems}, vol.~33, 2020.

\bibitem{shu2021agent}
T.~Shu, A.~Bhandwaldar, C.~Gan, K.~Smith, S.~Liu, D.~Gutfreund, E.~Spelke, J.~Tenenbaum, and T.~Ullman, ``Agent: A benchmark for core psychological reasoning,'' in \emph{International conference on machine learning}.\hskip 1em plus 0.5em minus 0.4em\relax PMLR, 2021, pp. 9614--9625.

\bibitem{jin2024mmtom}
C.~Jin, Y.~Wu, J.~Cao, J.~Xiang, Y.-L. Kuo, Z.~Hu, T.~Ullman, A.~Torralba, J.~B. Tenenbaum, and T.~Shu, ``Mmtom-qa: Multimodal theory of mind question answering,'' \emph{arXiv preprint arXiv:2401.08743}, 2024.

\bibitem{ying2024goma}
L.~Ying, K.~Jha, S.~Aarya, J.~B. Tenenbaum, A.~Torralba, and T.~Shu, ``Goma: Proactive embodied cooperative communication via goal-oriented mental alignment,'' \emph{arXiv preprint arXiv:2403.11075}, 2024.

\bibitem{puig2023nopa}
X.~Puig, T.~Shu, J.~B. Tenenbaum, and A.~Torralba, ``Nopa: Neurally-guided online probabilistic assistance for building socially intelligent home assistants,'' in \emph{2023 IEEE International Conference on Robotics and Automation (ICRA)}.\hskip 1em plus 0.5em minus 0.4em\relax IEEE, 2023, pp. 7628--7634.

\bibitem{devin2016implemented}
S.~Devin and R.~Alami, ``An implemented theory of mind to improve human-robot shared plans execution,'' in \emph{2016 11th ACM/IEEE International Conference on Human-Robot Interaction (HRI)}.\hskip 1em plus 0.5em minus 0.4em\relax IEEE, 2016, pp. 319--326.

\bibitem{hadfield2016cooperative}
D.~Hadfield-Menell, S.~J. Russell, P.~Abbeel, and A.~Dragan, ``Cooperative inverse reinforcement learning,'' in \emph{Advances in Neural Information Processing Systems}, 2016, pp. 3909--3917.

\bibitem{fisac2019pragmatic}
J.~F. Fisac, M.~A. Gates, J.~B. Hamrick, C.~Liu, D.~Hadfield-Menell, M.~Palaniappan, D.~Malik, S.~S. Sastry, T.~L. Griffiths, and A.~D. Dragan, ``Pragmatic-pedagogic value alignment,'' in \emph{Robotics research: the 18th international symposium Isrr}.\hskip 1em plus 0.5em minus 0.4em\relax Springer, 2019, pp. 49--57.

\bibitem{ying2025language}
L.~Ying, R.~Truong, K.~M. Collins, C.~E. Zhang, M.~Wei, T.~Brooke-Wilson, T.~Zhi-Xuan, L.~Wong, and J.~B. Tenenbaum, ``Language-informed synthesis of rational agent models for grounded theory-of-mind reasoning on-the-fly,'' \emph{arXiv preprint arXiv:2506.16755}, 2025.

\bibitem{reddy2019scalable}
C.~K. Reddy, E.~Beyrami, J.~Pool, R.~Cutler, S.~Srinivasan, and J.~Gehrke, ``A scalable noisy speech dataset and online subjective test framework,'' \emph{arXiv preprint arXiv:1909.08050}, 2019.

\bibitem{hart1968formal}
P.~E. Hart, N.~J. Nilsson, and B.~Raphael, ``A formal basis for the heuristic determination of minimum cost paths,'' \emph{IEEE transactions on Systems Science and Cybernetics}, vol.~4, no.~2, pp. 100--107, 1968.

\bibitem{fang2023anygrasp}
H.-S. Fang, C.~Wang, H.~Fang, M.~Gou, J.~Liu, H.~Yan, W.~Liu, Y.~Xie, and C.~Lu, ``Anygrasp: Robust and efficient grasp perception in spatial and temporal domains,'' \emph{IEEE Transactions on Robotics}, vol.~39, no.~5, pp. 3929--3945, 2023.

\end{thebibliography}
\bibliographystyle{IEEEtran}
\end{document}